\documentclass[letterpaper]{article} 
\usepackage{aaai25}  
\usepackage{times}  
\usepackage{helvet}  
\usepackage{courier}  
\usepackage[hyphens]{url}  
\usepackage{graphicx} 
\usepackage{natbib}  
\usepackage{caption} 
\usepackage{algorithm}
\usepackage{algorithmic}

\usepackage{newfloat}
\usepackage{listings}

\usepackage{subcaption} 
\usepackage{placeins} 
\usepackage{amsmath} 
\usepackage{amssymb} 
\usepackage{xcolor} 
\usepackage{multirow} 
\usepackage{booktabs} 
\usepackage{tabularx} 
\usepackage{pifont} 

\newcommand{\sys}[1]{\texttt{FedMoE}}

\DeclareCaptionStyle{ruled}{labelfont=normalfont,labelsep=colon,strut=off} 
\floatstyle{ruled}
\newfloat{listing}{tb}{lst}{}
\floatname{listing}{Listing}
\nocopyright 

\title{FedMoE: Personalized Federated Learning via Heterogeneous Mixture of Experts}
\author {
    Hanzi Mei\textsuperscript{\rm 1},
    Dongqi Cai\textsuperscript{\rm 1},
    Ao Zhou\textsuperscript{\rm 1},
    Shangguang Wang\textsuperscript{\rm 1},
    Mengwei Xu\textsuperscript{\rm 1}
}
\affiliations {
    \textsuperscript{\rm 1}Beijing University of Posts and Telecommunications\\
    \{meihanzi, cdq, aozhou, sgwang, mwx\}@bupt.edu.cn
}

\newcommand{\inlinecomment}[2][black]{\hfill \textit{\textcolor{#1}{// #2}}}

\begin{document}

\maketitle

\begin{abstract}
As Large Language Models (LLMs) push the boundaries of AI capabilities, their demand for data is growing. 
Much of this data is private and distributed across edge devices, making Federated Learning (FL) a de-facto alternative for fine-tuning (i.e., FedLLM).
However, it faces significant challenges due to the inherent heterogeneity among clients, including varying data distributions and diverse task types.
Towards a versatile FedLLM, we replace traditional dense model with a sparsely-activated Mixture-of-Experts (MoE) architecture, whose parallel feed-forward networks enable greater flexibility. 
To make it more practical in resource-constrained environments, we present \sys{}, the efficient personalized FL framework to address data heterogeneity, constructing an optimal sub-MoE for each client and bringing the knowledge back to global MoE. 
\sys{} is composed of two fine-tuning stages. 
In the first stage, \sys{} simplifies the problem by conducting a heuristic search based on observed activation patterns, which identifies a suboptimal submodel for each client.
In the second stage, these submodels are distributed to clients for further training and returned for server aggregating through a novel modular aggregation strategy. 
Meanwhile, \sys{} progressively adjusts the submodels to optimal through global expert recommendation. 
Experimental results demonstrate the superiority of our method over previous personalized FL methods.

\end{abstract}

%

\section{Introduction}
\begin{figure*}[t]
    \centering
    \begin{subfigure}[b]{0.19\textwidth}
        \includegraphics[width=\textwidth]{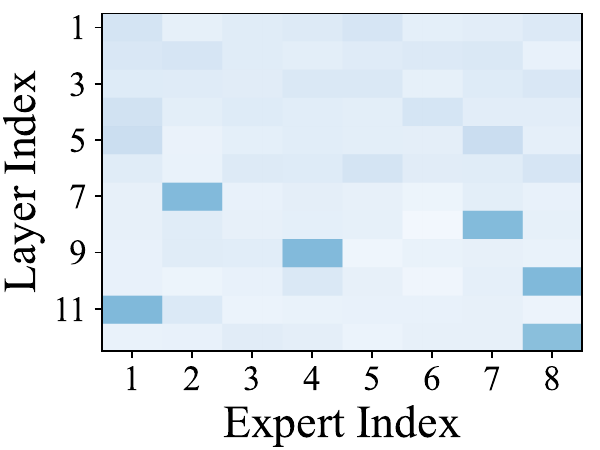}
        \captionsetup{justification=centering,singlelinecheck=false}
        \caption{Pretrained model on XSUM}
        \label{fig:pe-sub1}
    \end{subfigure}
    \begin{subfigure}[b]{0.19\textwidth}
        \includegraphics[width=\textwidth]{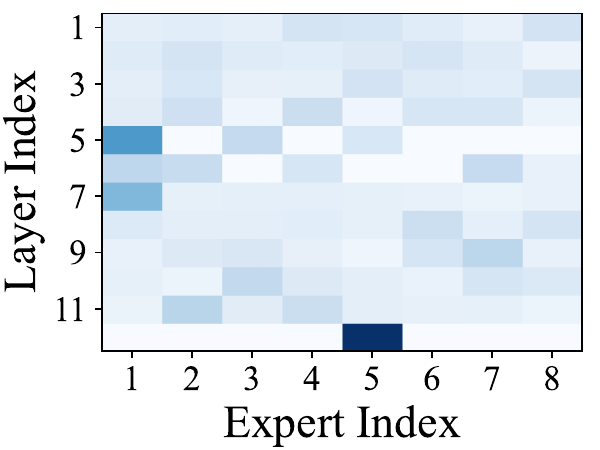}
        \captionsetup{justification=centering,singlelinecheck=false}
        \caption{5 Epochs fine-tuned on XSUM}
        \label{fig:pe-sub2}
    \end{subfigure}
    \begin{subfigure}[b]{0.19\textwidth}
        \includegraphics[width=\textwidth]{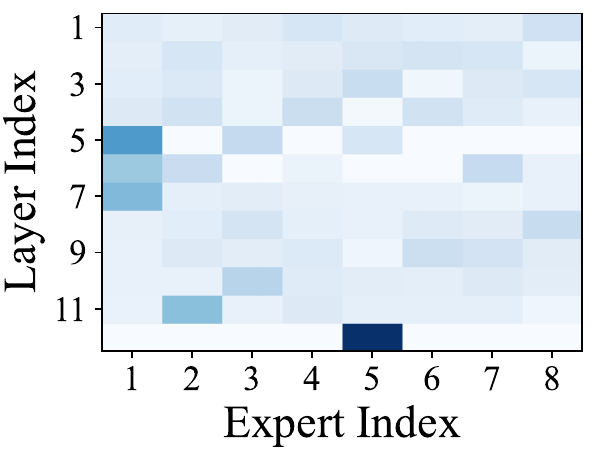}
        \captionsetup{justification=centering,singlelinecheck=false}
        \caption{20 Epochs fine-tuned on XSUM}
        \label{fig:pe-sub3}
    \end{subfigure}
    \begin{subfigure}[b]{0.19\textwidth}
        \includegraphics[width=\textwidth]{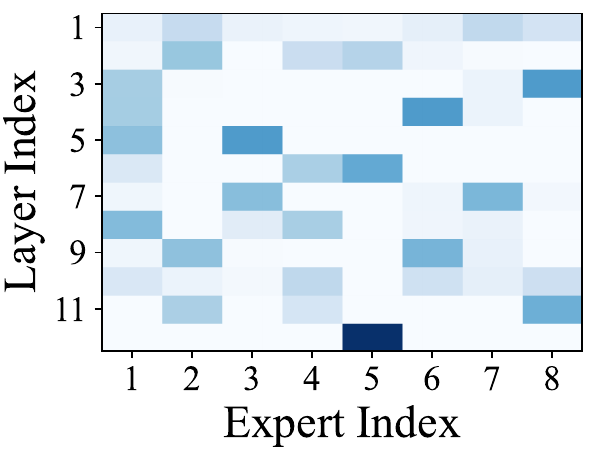}
        \captionsetup{justification=centering,singlelinecheck=false}
        \caption{20 Epochs fine-tuned on AG\_NEWS}
        \label{fig:pe-sub4}
    \end{subfigure}
    \begin{subfigure}[b]{0.19\textwidth}
        \includegraphics[width=\textwidth]{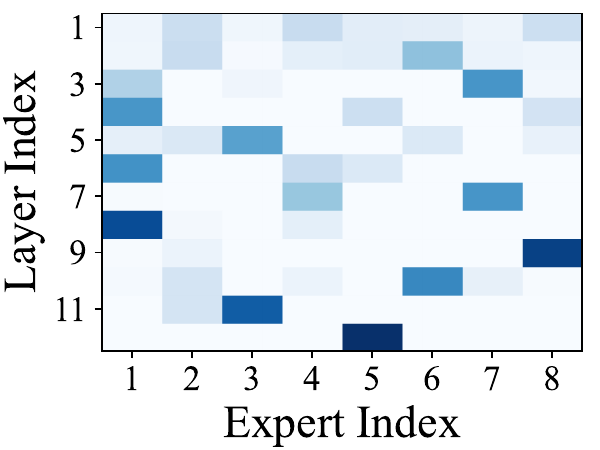}
        \captionsetup{justification=centering,singlelinecheck=false}
        \caption{20 Epochs fine-tuned on non-IID AG\_NEWS}
        \label{fig:pe-sub5}
    \end{subfigure}
    \caption{Heatmap of expert activation frequencies across different training stages and datasets.}
    \label{fig:Perliminary-experiments}
\end{figure*}
Emerging applications of Large Language Models (LLMs) like ChatGPT and Sora are transforming the AI landscape~\cite{wu2023brief, liu2024sora}. 
However, such models require increasing amounts of downstream data for fine-tuning~\cite{kaplan2020scaling}. 
Currently, the public cloud data is running out~\cite{villalobos2022will}. 
A huge amount of private data is still unexplored on distributed edge devices such as smartphones and autonomous vehicles. 
To leverage such distributed and private data, Federated Learning (FL) has become a de-facto approach~\cite{bonawitz2019towards}, allowing edge devices to process data on-device and collaboratively train stronger LLMs while preserving privacy, noted as FedLLM~\cite{xu2024fwdllm}.

Despite the promising vision, FedLLM remains challenging in heterogeneous reality. 
Generally, participants possess data that varies in domain and quality due to varying environments, typically noted as non-independent and identically distributed (non-IID) issue~\cite{zhu2021federated}. 
More critically, when extended to cross-task scenarios with different optimization objectives, the heterogeneity could be further exacerbated. 
This difficult but more realistic setting has been researched as task level~\cite{cai2023many,yao2022benchmark}. 
Therefore, personalized FL has emerged, with some approaches aiming to train proprietary and heterogeneous local models for each client.
Prior studies mainly focus on model distillation or model pruning derived from dense models~\cite{zhu2022resilient,ilhan2023scalefl}. 

Recently, Mixture-of-Experts (MoE)~\cite{rajbhandari2022deepspeed,jiang2024mixtral} has been widely explored to scale up the model efficiently, with experts pretrained to be sparsely activated for various tokens.
We thereby prototype \sys{}, the personalized FL framework that integrates transformer-based MoE into FL to address heterogeneity challenges. 
The impetus behind \sys{} stems from the unique and inherent features of MoE model. 
The expert-parallel structure along with the sparsely-activated mechanism has the potential to facilitate personalization for different clients, as each client flexibly activates the most relevant parameters. 
Furthermore, the MoE model supports on-demand scaling of model capacity~\cite{krajewski2024scaling}, enabling it to adapt to the increasing complexity of data and tasks in FL.

While the idea is intriguing, deploying MoE models into edge clients is not trivial. 
For example, training a Switch Transformers model~\cite{fedus2022switch} with 32 experts per layer demands at least 37GB of memory, overly exceeding the capacity of many client devices. 
To further facilitate the practical deployment of \sys{}, we propose an efficient \sys{} fine-tuning system to tackle data heterogeneity while considering resource constraints.
The main idea is to construct a specific, optimal sub-MoE model for each client, and then aggregate the knowledge back to the global MoE model. 
The holistic system is composed of two stages.
In the first stage, \sys{} efficiently reduces the problem complexity by identifying an initial submodel close to optimal for each client. Specifically, it conducts a heuristic search among all experts based on activation patterns observed during preliminary fine-tuning.
In the second stage, the initial submodels are sub-sampled from the global model and distributed to clients for edge training.
After edge training, \sys{} adopts a modular aggregation strategy in cloud server, sharing the relevant knowledge across clients while excluding negative interference. 
Additionally, \sys{} progressively adjusts the edge model structures through expert recommendation from the global perspective, striking the balance between efficiency and performance.

In general, our contributions are highlighted as follows:
\begin{itemize}
    \item We conduct preliminary experiments to demonstrate the characteristics of expert activation during heterogeneous FedLLM fine-tuning.
    \item We present \sys{}, an efficient FL system integrating transformer-based MoE models to address heterogeneity issues. 
    \sys{} dynamically searches and distributes personalized experts for different clients and absorbs the knowledge back into a generic global model.
    \item We conduct extensive experiments to demonstrate the empirical effectiveness of \sys{}. 
    Owing to meticulously designed personalization, \sys{} achieves superior performance across all tasks, while reducing the memory footprint and network traffic over existing baselines.
\end{itemize}
\section{Background and Motivation}
\subsection{Federated Learning}
The FL paradigm enables multiple edge devices to collaboratively train a shared model through rounds of edge-cloud communications, without sharing the local data~\cite{kairouz2021advances}. 
In each round, the cloud server randomly selects a set of edge devices (or \textit{clients}) and distributes the latest global model for edge fine-tuning. 
The server then collects those edge models to update the global model through aggregation. 
FL serves as a privacy-friendly training paradigm to utilize decentralized data under legal regulation such as GDPR~\cite{wiki:GDPR} and CCPA~\cite{wiki:CCPA}.

Despite FL's effectiveness in building privacy-preserving LLMs, training a generic large model across multiple tasks is still in its infancy. 
Given the divergent or even conflicting update directions produced by different tasks, it is challenging for global model to achieve optimal convergence~\cite{tan2022towards}. 
Furthermore, the fine-tuning process in FL is bottlenecked by on-device resources and network transmission~\cite{lim2020federated}. 
The global model must be affordable for the weakest device, which restricts the model scale and overall performance. Meanwhile, the resources of other devices are underutilized.
\subsection{Mixture of Experts}
Inspired by conditional computation, the MoE architecture breaks the principle of using the same set of parameters for all inputs~\cite{shazeer2017outrageously}. 
An MoE layer consists of a trainable gating network and a pool of experts, each a feed-forward network (FFN). The gating networks within the MoE determine the pathway of sequentially activated experts for each token. 
In the case of Switch Transformers, one of the most popular MoE-based models, the gating network is simplified to a top-1 router. Each token activates one expert per layer based on the probabilities computed by router, leaving the other experts unupdated.

MoE architecture scales up the model capacity in a computationally efficient manner, achieving significant performance improvements across many downstream tasks~\cite{fedus2022review}. 
The effectiveness also extends from individual tasks to multi-task learning scenarios, where multiple objectives are learned simultaneously~\cite{gupta2022sparsely, kim2021scalable}.
The advantage of MoE can be attributed to the unique expert-parallel structure and sparsely-activated mechanism, rendering each expert capable of handling specific tasks or data subsets. Therefore, MoE model shows significant potential for adapting to multi-tasking and heterogeneous data environments.
\subsection{Motivation and Preliminary Experiments}
Sharing all parameters across clients is not a silver-bullet approach in FL, especially when dealing with heterogeneous data. 
Unlike previous FL studies, we aim to train an MoE-architectured backbone. The parameter spaces in sparse layers are disentangled, allowing for a more flexible and on-demand parameter-sharing scheme.
Specifically, a subset of optimal experts can be sub-sampled from the global model, constructing a client-specific heterogeneous model for personalized learning, and bringing the knowledge back to the global model during aggregation. 
This novel scheme allows clients to share the most relevant parameters, encouraging cooperation while preventing interference.

However, the main obstacle lies in identifying the optimal experts for specific data under the resource constraints. To gain insights for the design of \sys{}, we conduct a series of preliminary experiments based on Switch Transformers to explore the characteristics of expert activation.

\textbf{Observation-1: Expert activation characteristics are skewed and dynamic during fine-tuning.} As illustrated in Figure~\ref{fig:Perliminary-experiments}, experts exhibit a distinct imbalance in activation frequencies, indicating that only certain experts (those frequently activated) are of critical importance~\cite{yi2023edgemoe,lu2024not}. The comparison between Figure~\ref{fig:pe-sub1} and Figure~\ref{fig:pe-sub3} further reveals that fine-tuning significantly alters the activation patterns, enabling the model to identify the optimal experts progressively. This observation suggests that the importance of experts should be assessed dynamically during training, rather than statically beforehand.

\begin{figure}[t]
    \centering
    \begin{subfigure}[t]{0.22\textwidth}
        \includegraphics[width=\textwidth]{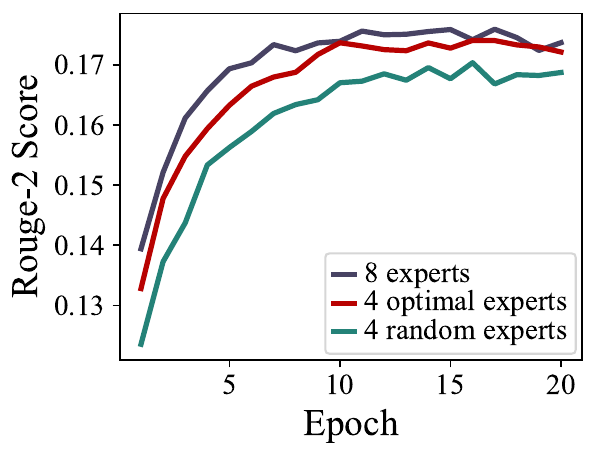}
        \caption{Accuracy}
        \label{fig:sub1}
    \end{subfigure}
    \begin{subfigure}[t]{0.22\textwidth}
        \includegraphics[width=\textwidth]{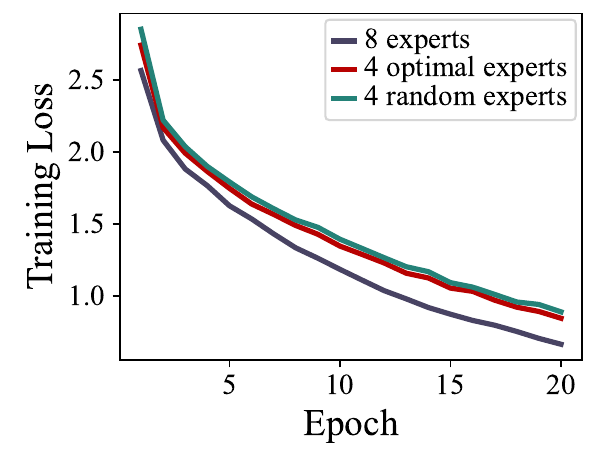}
        \caption{Loss}
        \label{fig:sub2}
    \end{subfigure}
    \caption{Performance of full model (8 experts per layer) and submodels (4 optimal experts per layer, 4 random experts per layer).}
    \label{fig:Perliminary-experiments-2}
\end{figure}

\textbf{Observation-2: Optimal experts converge quickly and perform comparably.} We monitor the activation characteristics at every epoch during fine-tuning on a text summarization task, whose evolution is shown in Figure~\ref{fig:pe-sub1}, Figure~\ref{fig:pe-sub2} and Figure~\ref{fig:pe-sub3}. The optimal experts quickly stabilize after a certain epoch, exhibiting only minor fluctuations in their activation frequencies, which facilitates model pruning. As demonstrated in Figure~\ref{fig:Perliminary-experiments-2}, the submodel constructed by optimal experts shows a comparable performance to full model, while the submodel built with randomly chosen experts falls short.

\textbf{Observation-3: Optimal experts differ across heterogeneous data, especially when heterogeneity reaches the task level.} As illustrated in Figure~\ref{fig:pe-sub4} and Figure~\ref{fig:pe-sub5}, within the same task, data of diverse distributions favor different subsets of experts, albeit with some overlap. The difference is exacerbated when it comes to different tasks that require distinct knowledge, as demonstrated by Figure~\ref{fig:pe-sub3} and Figure~\ref{fig:pe-sub5}. 
\section{Method}
\subsection{Problem Formulation}
In personalized FL, there are $n$ clients collaborating to learn $m$ types of downstream tasks under varying memory constraints. Each client $k = 1,2, \ldots ,n$ executes edge training based on a local dataset ${\mathcal{D}_k}: = \left( {\mathcal{X},{\mathcal{Y}_k},{p_k}\left( x \right)} \right)$. The input space $\mathcal{X}$ is globally shared, while the input distribution $p_k(x)$ can be either IID or non-IID across different clients. The output space $\mathcal{Y}_k$ corresponds to a target task $\mathcal{T}_k \in \left\{ {1,2, \ldots ,m} \right\}$. The union $\mathcal{D} = \bigcup\nolimits_{k = 1}^n {{\mathcal{D}_k}}$ represents the overall dataset used for FL.

The local objective of client $k$ is to minimize the sum of label-smoothed cross-entropy loss and the weighted load balance loss for MoE, while keeping the memory usage of the edge model $w_k$ within the memory limit $M_k$, formulated as:
\begin{gather}
    \min_{w_k} \, \mathcal{L}_k = \mathbb{E}_{(x,y) \sim \mathcal{D}_k}\left[ \mathcal{L}_{CE}(y,\hat{y}(x;w_k)) + \alpha \mathcal{L}_{LB} \right] \notag \\
    \text{s.t.} \quad \text{mem}(w_k) \leq M_k
\end{gather}
The clients in FL collaboratively train a global model $w$ to minimize the global objective function, formulated as:
\begin{gather}
    \min_{w} \, \mathcal{L} = \sum\limits_{k = 1}^n {\frac{{\left| {{\mathcal{D}_k}} \right|}}{{\left| \mathcal{D} \right|}}} {\mathcal{L}_k}\left( {{w_k}} \right)
\end{gather}
\subsection{Overview}
\subsubsection{Model Structure}
The cloud hosts a large MoE model with an equal number of experts at every layer, initialized by a pre-trained MoE model. Global model's extensive capacity enables it to provide a broad range of knowledge and store new information effectively. 
The clients host heterogeneous sub-MoEs, differing in both model architecture (i.e., different expert numbers per layer) and parameter space (e.g., different pieces of the global model). These client-specific submodels are sub-sampled from the global model, retaining the most relevant experts to flexibly adapt to data characteristics under memory constraints. 

\subsubsection{Workflow}
\begin{figure*}[!t]
    \centering
    \includegraphics[width=2.1\columnwidth]{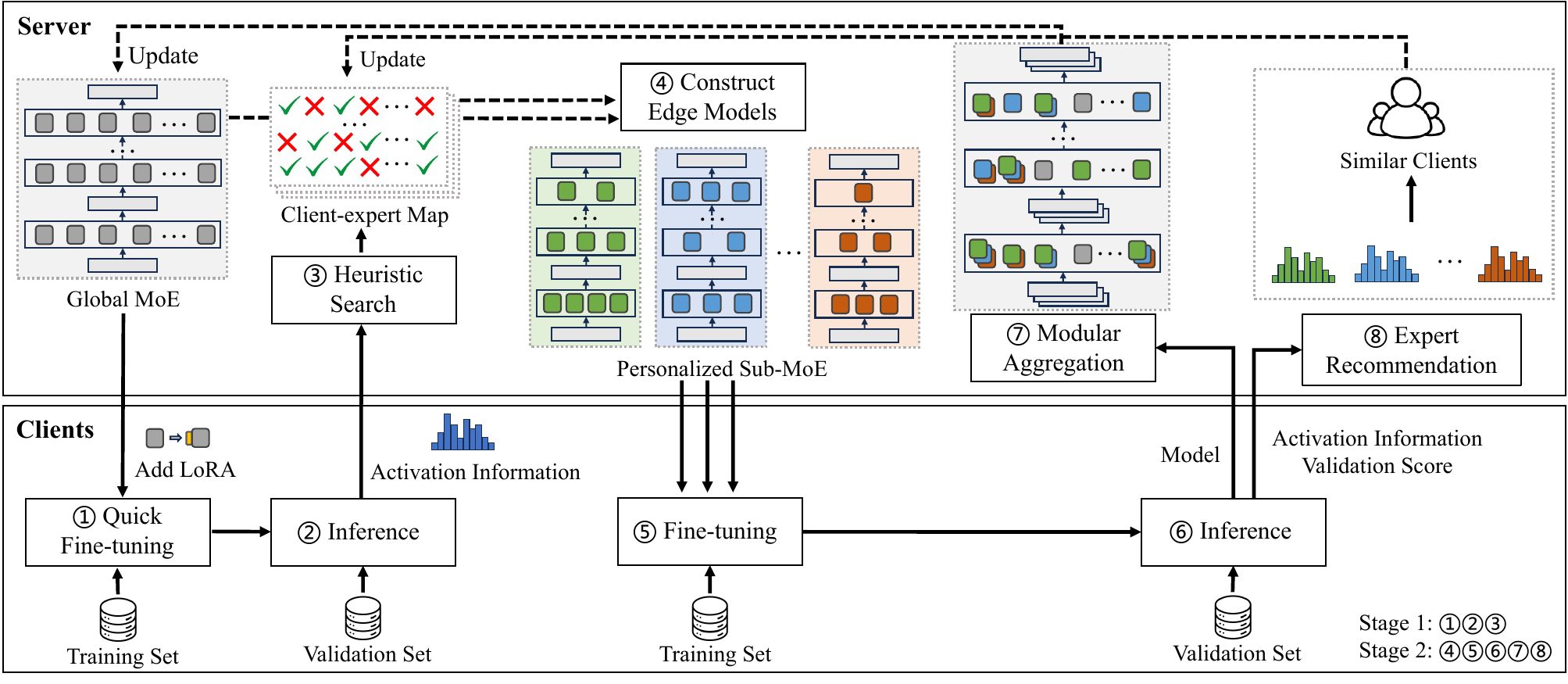}
    \caption{FedMoE workflow.}
    \label{fig:training-workflow}
\end{figure*}

\sys{} is structured into two stages, as illustrated in Figure~\ref{fig:training-workflow}. 
In the first stage, the server collects activation information from clients after memory-efficient fine-tuning at the edge (steps \ding{172}–\ding{173}). Based on this information, the server conducts coarse-grained submodel searches (step \ding{174}) to determine an approximate model architecture for each client (i.e., an initial mapping of each client to its preferred expert subset). In the second stage, federated learning begins with the submodels initialized in stage one (step \ding{175}). In each round, fine-tuned submodels from different clients are integrated into the global model through modular aggregation (step \ding{176}–\ding{178}), followed by fine-grained structural adjustments guided by real-time feedback (step \ding{179}).
\subsection{Stage One: Coarse-grained Submodel Initialization}

To begin with, we employ expert activation probability as a criterion to assess expert importance under specific data distributions, represented as $p_{i,j} = \frac{{{n_{i,j}}}}{N}$, where ${n_{i,j}}$ is the activation times of the $j$-th expert in $i$-th layer and $N$ is the total token count. 
Experts with higher activation probabilities are more crucial for the downstream task. Stage one involves a large-scale problem, as it requires fairly evaluating the importance of all experts across a vast number of clients. Conducting such evaluation within the memory limits of edge devices exacerbates the complexity.

\subsubsection{Activation Probability Collection}

The cloud sends a full-scale MoE to clients with all experts working in parallel.
Clients with sufficient memory further equip each expert with LoRA structure and then perform Parameter-Efficient Fine-Tuning (PEFT).
PEFT is a well-established method for memory-efficient fine-tuning~\cite{hu2021lora,sun2022recent}.
Since expert preferences converge quickly, fine-tuning only lasts a few rounds (e.g., 5 rounds) to obtain expert activation probabilities on validation datasets.
After gathering activation information, the cloud server performs a static prediction of expert activation probabilities for memory-insufficient clients, based on the intuition that homogeneous data typically favor a similar set of experts.
Specifically, the cloud calculates the estimated probabilities for a given client by averaging the probabilities of other clients with the same task, weighted by their data volume.
Ultimately, the cloud collects activation probabilities for all experts across all clients in a rough yet efficient manner, laying the foundation for constructing personalized models.

\subsubsection{Heuristic Submodel Search}

Based on the activation probabilities, the cloud strives to initialize a suboptimal submodel for each client under its memory constraint.
The issue is further modeled as an optimization problem.
For each client, the cloud aims to find a maximum threshold $\theta$, ensuring the combined activation probabilities of the experts retained in each layer are at least $\theta$. The uniform threshold across each layer guarantees the optimal performance of this model, as some layers only require a few skilled experts, whereas others demand more to collaborate. Additionally, the memory usage of the retained experts and dense layers must not exceed $\alpha (0 < \alpha  \leqslant 1)$ of total available memory. The reserved memory allows for fine-grained model structural adjustments in the second stage, which potentially requires more experts. Solving for the binary variable ${x_{i,j}} \in \left\{ {0,1} \right\}$ that indicates whether to retain the $j$-th expert in the $i$-th layer, the problem is formulated as:
\begin{gather}
    \max_{x_{i,j}} \, \theta \notag \\
    \text{s.t.} \quad \left\{ {\begin{array}{*{20}{c}}
        {\sum\limits_{j = 1}^{{E_i}} {{x_{i,j}}{p_{i,j}} \geqslant \theta } ,\forall i = 1, \ldots ,L} \\ 
        {\text{mem}\left( {\sum\limits_{i = 1}^L {\sum\limits_{j = 1}^{{E_i}} {{x_{i,j}}{w_{i,j}} + {w_d}} } } \right) \leqslant \alpha \cdot M} 
      \end{array}} \right.
\end{gather}
Since solving the NP-hard multi-dimensional knapsack problem directly is computationally prohibitive, we instead employ a heuristic algorithm based on binary search.
In practice, the cloud searches for the optimal threshold $\theta$ within the range of $\left[ {0,1} \right]$ by gradually adjusting the bounds of feasible region. For a given threshold, \sys{} attempts to construct a smallest submodel with respect to the threshold and verify whether it exceeds the memory limit. 
If not exceed, the lower bound of $\theta$ is adjusted upward to find a more effective submodel; otherwise, the upper bound of $\theta$ is adjusted downward to alleviate memory consumption. The efficient search initializes the client-expert map in the cloud, achieving an ideal balance between memory usage and performance.

\begin{algorithm}[t]
\caption{Federated training and submodel adjustments in the second stage.}
\label{alg:fedmoe-stage2}
\begin{algorithmic}[1] 
\FOR{each round $r = 1, \ldots, R$}
    \STATE $\mathcal{S} \leftarrow$ sample subset of clients $\mathcal{U}=\{u_1, \ldots, u_n\}$
    \STATE Construct and send $w_k$ to each $u_k \in \mathcal{S}$ 
    \FOR{client $u_k \in \mathcal{S}$ in parallel}
        \STATE $w_k^* \leftarrow \textsc{train}(w_k, \mathcal{D}_k^{\text{train}})$
        \STATE $p_{\text{all\_experts}}, \mathit{acc} \leftarrow \textsc{validate}(w_k^*, \mathcal{D}_k^{\text{val}})$
        \STATE send $w_k^*, p_{\text{all\_experts}}, \mathrm{acc}$ to server
    \ENDFOR
    \STATE $w_{global} \leftarrow$ modular aggregate $w_k^*$ for $u_k$ in $\mathcal{S}$
    \FOR{client $u_k \in \mathcal{S}$}
        \IF{$\mathit{acc}$ not improved}
            \FOR{client $u_a \in \mathcal{U} \setminus {u_k}$}
                \STATE Caculate $\text{sim}\left( {{u_k},{u_a}} \right)$ \inlinecomment{refer to equation~\eqref{eq:similarity}}
            \ENDFOR
            \STATE $\mathcal{S}' \leftarrow$ top $K$ similar clients
            \STATE $n = \textsc{avg}(n_{\text{expert}}(\mathcal{S}')) - n_{\text{expert}}(u_k)$
            \IF{$n > 0$}
                \FOR{expert out of $w_k$}
                    \STATE caculate $\hat{p}_{\text{expert}}$ \inlinecomment{refer to equation~\eqref{eq:estimated probability}}
                \ENDFOR
                \STATE $\mathcal{E} \leftarrow$ top $n$ experts ranked by highest $\hat{p}_{\text{expert}}$
                \STATE add experts $\mathcal{E}$ to submodel $w_k$
            \ELSE
                \FOR{expert within $w_k$}
                    \STATE caculate $\hat{p}_{\text{expert}}$ \inlinecomment{refer to equation~\eqref{eq:estimated probability}}
                \ENDFOR
                \STATE $\mathcal{E} \leftarrow$ top $n$ experts ranked by lowest $\hat{p}_{\text{expert}}$
                \STATE remove experts $\mathcal{E}$ out of submodel $w_k$
            \ENDIF
        \ENDIF
    \ENDFOR
\ENDFOR
\end{algorithmic}
\end{algorithm}
\subsection{Stage Two: Federated Training and Fine-grained Submodel Adjustment}
In the second stage shown in Algorithm~\ref{alg:fedmoe-stage2}, federated training begins with the submodels initialized by the first stage. After fine-tuning on clients' private data, the submodel is filled with personalized knowledge. The challenge lies in how to encourage cooperation while reducing interference among heterogeneous submodels during aggregation. Additionally, initial submodels may prove suboptimal, either failing to capture personalized knowledge (\textit{underfitting}) or being excessively redundant (\textit{overfitting}). The cloud serves as a central node, crucial not only for managing model parameters but also for structural model adjustments. Specifically, \sys{} incorporates clients with diverse data distributions and various task types through modular model sub-sampling and aggregation. Afterwards, the cloud progressively adjusts the submodels based on the expert recommendation mechanism from a global perspective, continually optimizing the overall system.

\subsubsection{Submodel Deployment}

At the beginning of each round, a random selection of clients is chosen to participate in the training. The cloud sub-samples the dense layers and optimal expert subset from the global MoE based on the latest client-expert map, constructing a personalized submodel for each client. Deployed on edge devices, these submodels are fine-tuned on local training dataset. Subsequently, clients perform inference on their validation dataset, collecting the up-to-date expert activation probabilities and validation scores which are sent to cloud along with the edge model parameters.

\subsubsection{Modular Aggregation}

After receiving all edge models, the cloud integrates newly-learned knowledge back into the global model based on the modular aggregation strategy. For parameters of dense layers, the cloud conventionally employs the Federated Averaging (FedAvg) strategy. For parameters of sparse layers, aggregation becomes more complicated due to the heterogeneous expert subsets across clients. In detail, unactivated experts in the global model remain unchanged, experts used by a single client are updated directly, and experts shared by multiple clients are aggregated based on the FedAvg strategy. The expert-corresponding dimensions of routers are updated in the same pattern. Module-granular updates allow for independent optimization of different parts within the global model, fostering collaboration and mutual enhancement while preventing conflicts across different clients.

\subsubsection{Expert Recommendation}

Besides updating the global model, the cloud precisely optimizes the submodel structure to ensure learning efficiency throughout the training cycle. If a client shows no performance gains after several rounds, its model is considered to have reached a bottleneck. 
The cloud leverages insights provided by other clients from a global perspective, conducting expert-granular structural adjustments. Specifically, the cloud calculates the cosine similarity between the current client and other clients based on their expert activation probabilities, formulated as:
\begin{gather}
    \label{eq:similarity}
    \text{sim}\left( {{\mathbf{u}_k},{\mathbf{u}_a}} \right) = \frac{{\sum\limits_{i = 1}^L {\sum\limits_{j = 1}^{{E_i}} {\left( {{{\left( {{p_{i,j}}} \right)}_{{\mathbf{u}_k}}} \times {{\left( {{p_{i,j}}} \right)}_{{\mathbf{u}_a}}}} \right)} } }}{{\sqrt {\sum\limits_{i = 1}^L {\sum\limits_{j = 1}^{{E_i}} {{{\left( {{p_{i,j}}} \right)}_{{\mathbf{u}_k}}}^2} } }  \times \sqrt {\sum\limits_{i = 1}^L {\sum\limits_{j = 1}^{{E_i}} {{{\left( {{p_{i,j}}} \right)}_{{\mathbf{u}_a}}}^2} } } }}
\end{gather}
The current client takes the top $K$ clients ranked by similarity as reference, as indicated by:
\begin{gather}
    \mathbb{S}\left( {{\mathbf{u}_k}} \right) = \left\{ {\left. {{\mathbf{u}_a}} \right|rank \, \text{sim}\left( {{\mathbf{u}_k},{\mathbf{u}_a}} \right) \leqslant K,a \ne k} \right\}
\end{gather}
If their number of experts exceeds that of the current client, the cloud recommends incorporating more effective experts into its submodel; otherwise, it suggests pruning underperforming experts. To fairly evaluate the effectiveness of all experts (including those within and outside the submodel), the cloud estimates their activation probabilities based on those of top-k most similar clients, formulated as:
\begin{gather}
    \label{eq:estimated probability}
    {\left( {{{\hat p}_{i,j}}} \right)_{{\mathbf{u}_k}}} = \frac{{\sum\limits_{{\mathbf{u}_a} \in \mathbb{S}\left( {{\mathbf{u}_k}} \right)} {\text{sim}\left( {{\mathbf{u}_k},{\mathbf{u}_a}} \right) \times {{\left( {{p_{i,j}}} \right)}_{{\mathbf{u}_a}}}} }}{{\sum\limits_{{\mathbf{u}_a} \in \mathbb{S}\left( {{\mathbf{u}_k}} \right)} {\text{sim}\left( {{\mathbf{u}_k},{\mathbf{u}_a}} \right)} }}
\end{gather}
Note that this is an exploratory adjustment. If the performance of the adjusted submodel does not improve, its structure will revert to the previous version and remain fixed thereafter.
\section{Experiments}
\begin{table*}[t]
\centering
\caption{End-to-end performance of different personalized methods in various FL settings.}
\label{tab:end-to-end}
\begin{subtable}{.5\linewidth}
\centering
\caption{Setting: Standard-Hetero-T}
\label{subtab:setting1}
\begin{tabular}{@{}>{\centering\arraybackslash}p{2.3cm}%
    >{\centering\arraybackslash}p{0.5cm}%
    >{\centering\arraybackslash}p{0.5cm}%
    >{\centering\arraybackslash}p{0.5cm}%
    >{\centering\arraybackslash}p{1.1cm}%
    >{\centering\arraybackslash}p{1.6cm}@{}}
\toprule
\multirow{3}{*}{Method} & \multicolumn{3}{c}{Performance} & \multirow{3}{*}{\begin{tabular}[c]{@{}c@{}}Comm.\\ Vol. (GB)\end{tabular}} & \multirow{3}{*}{\begin{tabular}[c]{@{}c@{}}Mem.\\Usage (GB)\end{tabular}} \\
                        & task-TC    & task-RC    & task-TS    &                                                                                      &                                                                             \\
\midrule
randomMoE               & 91.63           & 84.23           & 14.51           & 2.30            & 15.63        \\
FedProx                 & 92.92           & \textbf{87.99}  & 11.94           & 2.30            & 24.71         \\
SCAFFOLD                & 85.98           & 69.44           & 5.86            & 4.61            & 17.29          \\
FedMoE (Ours)            & \textbf{94.76}  & 86.64           & \textbf{16.92}  & \textbf{1.76}   & \textbf{13.44} \\
\bottomrule
\end{tabular}
\end{subtable}%
\begin{subtable}{.5\linewidth}
\centering
\caption{Setting: Standard-Hetero-TD}
\label{subtab:setting2}
\begin{tabular}{@{}>{\centering\arraybackslash}p{2.3cm}%
    >{\centering\arraybackslash}p{0.5cm}%
    >{\centering\arraybackslash}p{0.5cm}%
    >{\centering\arraybackslash}p{0.5cm}%
    >{\centering\arraybackslash}p{1.1cm}%
    >{\centering\arraybackslash}p{1.6cm}@{}}
\toprule
\multirow{3}{*}{Method} & \multicolumn{3}{c}{Performance} & \multirow{3}{*}{\begin{tabular}[c]{@{}c@{}}Comm.\\ Vol. (GB)\end{tabular}} & \multirow{3}{*}{\begin{tabular}[c]{@{}c@{}}Mem.\\Usage (GB)\end{tabular}} \\
                        & task-TC    & task-RC    & task-TS    &                                                                                      &                                                                             \\
\midrule
randomMoE               & 34.19           & 82.93           & 13.51           & 2.30           & 15.56            \\
FedProx                 & 85.09           & \textbf{87.57}  & 11.76           & 2.30           & 24.63            \\
SCAFFOLD                & 61.72           & 67.17           & 5.83            & 4.61           & 17.23             \\
FedMoE (Ours)            & \textbf{88.44}  & 86.55           & \textbf{16.63}  & \textbf{1.85}  & \textbf{13.89}      \\
\bottomrule
\end{tabular}
\end{subtable}%

\begin{subtable}{.5\linewidth}
\centering
\caption{Setting: Enforced-Hetero-T}
\label{subtab:setting3}
\begin{tabular}{@{}>{\centering\arraybackslash}p{2.3cm}%
    >{\centering\arraybackslash}p{0.5cm}%
    >{\centering\arraybackslash}p{0.5cm}%
    >{\centering\arraybackslash}p{0.5cm}%
    >{\centering\arraybackslash}p{1.1cm}%
    >{\centering\arraybackslash}p{1.6cm}@{}}
\toprule
\multirow{3}{*}{Method} & \multicolumn{3}{c}{Performance} & \multirow{3}{*}{\begin{tabular}[c]{@{}c@{}}Comm.\\ Vol. (GB)\end{tabular}} & \multirow{3}{*}{\begin{tabular}[c]{@{}c@{}}Mem.\\Usage (GB)\end{tabular}} \\
                        & task-TC    & task-RC    & task-TS    &                                                                                      &                                                                             \\
\midrule
randomMoE               & 88.86           & 82.68           & 14.37           & 2.30           & 15.44            \\
FedProx                 & 92.51           & \textbf{86.69}  & 11.88           & 2.30           & 24.55            \\
SCAFFOLD                & 36.17           & 72.51           & 6.78            & 4.61           & 17.21            \\
FedMoE (Ours)            & \textbf{94.85}  & 85.79           & \textbf{17.19}  & \textbf{1.89}  & \textbf{13.78}    \\
\bottomrule
\end{tabular}
\end{subtable}%
\begin{subtable}{.5\linewidth}
\centering
\caption{Setting: Enforced-Hetero-TD}
\label{subtab:setting4}
\begin{tabular}{@{}>{\centering\arraybackslash}p{2.3cm}%
    >{\centering\arraybackslash}p{0.5cm}%
    >{\centering\arraybackslash}p{0.5cm}%
    >{\centering\arraybackslash}p{0.5cm}%
    >{\centering\arraybackslash}p{1.1cm}%
    >{\centering\arraybackslash}p{1.6cm}@{}}
\toprule
\multirow{3}{*}{Method} & \multicolumn{3}{c}{Performance} & \multirow{3}{*}{\begin{tabular}[c]{@{}c@{}}Comm.\\ Vol. (GB)\end{tabular}} & \multirow{3}{*}{\begin{tabular}[c]{@{}c@{}}Mem.\\Usage (GB)\end{tabular}} \\
                        & task-TC    & task-RC    & task-TS    &                                                                                      &                                                                             \\
\midrule
randomMoE               & 37.81           & 82.61           & 13.80           & 2.30           & 15.45            \\
FedProx                 & \textbf{73.85}  & \textbf{86.59}  & 12.03           & 2.30           & 24.51            \\
SCAFFOLD                & 69.34           & 70.69           & 6.96            & 4.61           & 17.18            \\
FedMoE (Ours)            & 73.56           & 85.84           & \textbf{16.78}  & \textbf{1.95}  & \textbf{14.04}    \\
\bottomrule
\end{tabular}
\end{subtable}%

\end{table*}

\subsection{Experimental Setups}
\subsubsection{FL Simulations}
We create four FL simulations based on three classic NLP downstream datasets, including AG News~\cite{zhang2015character} for text classification (task-TC), SQuAD~\cite{rajpurkar2016squad} for reading comprehension (task-RC) and XSum~\cite{narayan2018don} for text summarization (task-TS). The evaluation metrics are accuracy, F1 score and Rouge-2 score, respectively. Each simulation setting mimics a complex real-world scenario. (1) \textbf{Standard-Hetero-T} involves 30 clients with heterogeneous tasks, and 5 clients are randomly selected for training each round. (2) \textbf{Standard-Hetero-TD} further introduces data heterogeneity by assigning label-skewed non-IID datasets to clients with identical tasks, following the methodology in FedNLP~\cite{lin2021fednlp}. (3) \textbf{Enforced-Hetero-T} simulates a more conflicting scenario by forcibly selecting 3 clients with different task types in each training round, using the same 30-client setup as in Standard-Hetero-T. (4) \textbf{Enforced-Hetero-TD} adopts conflicting client selection scheme and uses the same 30-client setup as in Standard-Hetero-TD. The clients in the above settings are resource-heterogeneous, with memory capacities ranging from 18GB to 24GB, typical of high-performance smartphones and edge computing platforms.

\subsubsection{Baselines}
We compare \sys{} with three personalized FL methods: (1) \textbf{randomMoE} constructs personalized edge models by randomly selecting a subset of experts from the global MoE, ensuring a degree of information isolation among clients. (2) \textbf{FedProx}~\cite{li2020federated} is a federated optimization algorithm that incorporates a regularization term during local updates to mitigate the impacts of heterogeneity. (3) \textbf{SCAFFOLD}~\cite{karimireddy2020scaffold} employs control variates to correct the local update directions, overcoming client-drift caused by heterogeneous data.

\subsubsection{Models}
The following experiments are conducted based on Switch Transformers architecture whose pre-trained weights are downloaded directly from Hugging Face~\cite{wolf2019huggingface}. The global models for \sys{} and randomMoE are configured with 32 experts per layer, whereas others are configured with 8 experts per layer due to edge device memory constraints.

\subsection{End-to-End Performance}
\begin{figure*}[t]
    \centering
    \begin{subfigure}[t]{0.19\textwidth}
        \includegraphics[width=\textwidth]{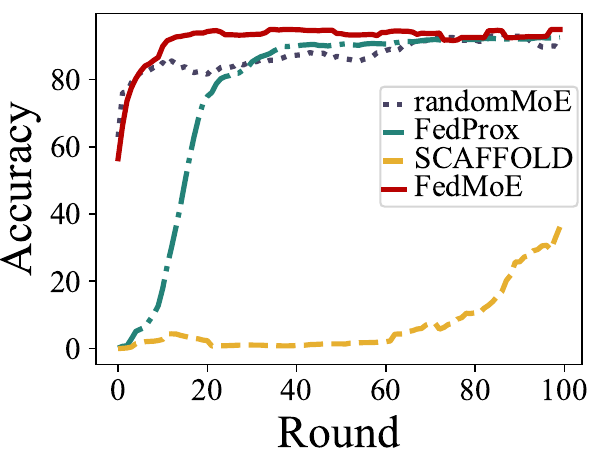}
        \captionsetup{justification=centering,singlelinecheck=false}
        \caption{Performance on task-TC}
        \label{fig:ete-sub1}
    \end{subfigure}
    \begin{subfigure}[t]{0.19\textwidth}
        \includegraphics[width=\textwidth]{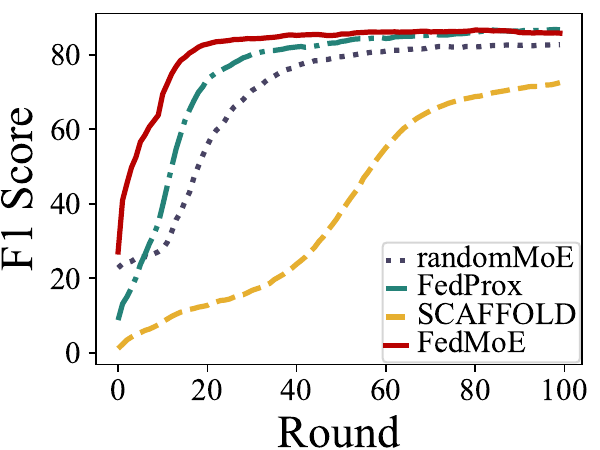}
        \captionsetup{justification=centering,singlelinecheck=false}
        \caption{Performance on task-RC}
        \label{fig:ete-sub2}
    \end{subfigure}
    \begin{subfigure}[t]{0.19\textwidth}
        \includegraphics[width=\textwidth]{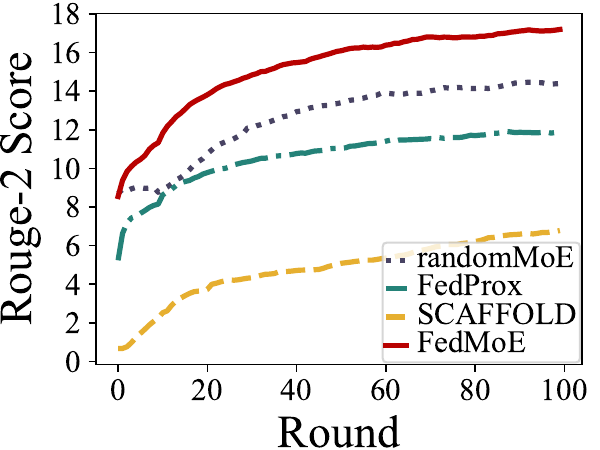}
        \captionsetup{justification=centering,singlelinecheck=false}
        \caption{Performance on task-TS}
        \label{fig:ete-sub3}
    \end{subfigure}
    \begin{subfigure}[t]{0.19\textwidth}
        \includegraphics[width=\textwidth]{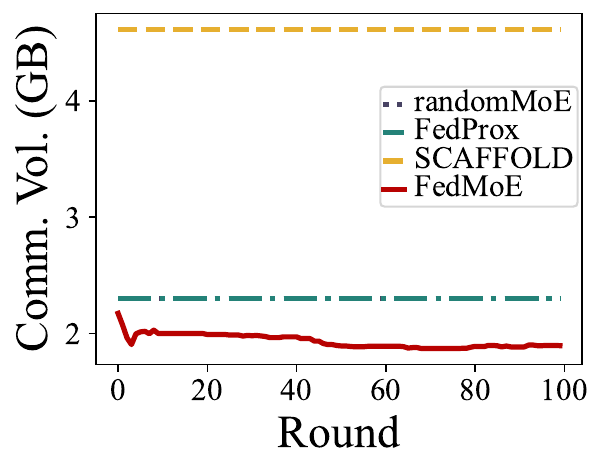}
        \captionsetup{justification=centering,singlelinecheck=false}
        \caption{Communication volume}
        \label{fig:ete-sub4}
    \end{subfigure}
    \begin{subfigure}[t]{0.19\textwidth}
        \includegraphics[width=\textwidth]{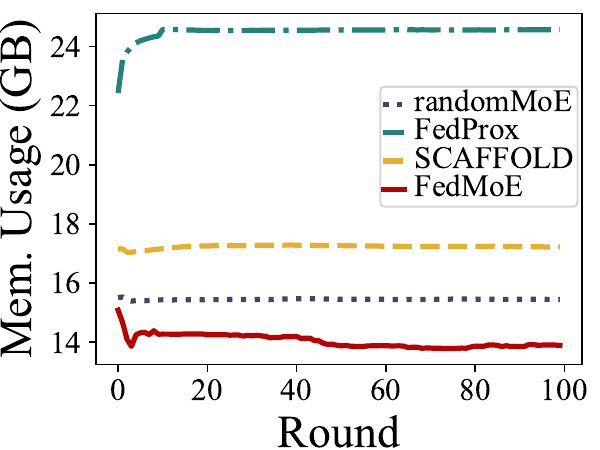}
        \captionsetup{justification=centering,singlelinecheck=false}
        \caption{Peak memory usage}
        \label{fig:ete-sub5}
    \end{subfigure}
    \caption{End-to-end performance of different personalized methods during training under Enforced-Hetero-T setting.}
    \label{fig:end-to-end performance}
\end{figure*}

Table~\ref{tab:end-to-end} summarizes end-to-end performance of four methods in various FL settings, including task performance, communication volume and peak memory usage. Figure~\ref{fig:end-to-end performance} shows the end-to-end performance throughout the training process in one of the settings.
From the results, we draw four key observations as follows:

\textbf{FedMoE outperforms baselines in overall task performance.}
\sys{} typically demonstrates comparable or superior performance across various tasks, achieving average improvements of 7.63\%, 15.50\%, 10.00\%, and 12.74\% over the best-performing baseline  in the four settings. Notably, for complex tasks such as task-TS, \sys{} beats the best baseline non-trivially with improvements of up to 23.09\%, effectively avoiding the knowledge interference between different tasks that occurs in FedProx and SCAFFOLD.

\textbf{FedMoE achieves faster convergence.}
Figures~\ref{fig:ete-sub1}--\ref{fig:ete-sub3} illustrate the convergence process across various task types in Enforced-Hetero-T Setting. To achieve 99\% of the relative target performance, \sys{} demonstrates a speedup of 1.65×, 2.13×, and 1.64× compared to the three baselines, and 1.35×, 2.06×, and 2.92× for 90\% of the target performance. 

\textbf{FedMoE reduces communication overhead and memory usage significantly, making personalized FL resource-efficient.}
The communication overhead of \sys{} decreases by 19.11\%, 19.11\%, and 59.56\% on average compared to the three baselines, and memory consumption decreases by 11.16\%, 43.95\%, and 19.97\%, respectively. 
The optimization methods in FedProx and SCAFFOLD introduce additional memory and communication overheads, which are accumulated during long-term FL training. In contrast, the cost of FedMoE is negligible, with only a one-time roundtrip communication of about 7.46GB and a memory overhead of approximately 13.06GB that lasts for only a few rounds in the first stage.

\textbf{FedMoE demonstrates strong robustness and stable performance in complex scenarios.}
We calculate the Coefficient of Variation (CV), denoted as ${c_v} = \frac{\sigma }{\mu }$, across various settings for each task and use the average value as the Composite Variation Index (CVI) to assess the robustness of each method. \sys{} with a notably small CVI of 0.0445 and FedProx with a CVI of 0.0569 exhibit relatively stable performance, whereas randomMoE with a CVI of 0.1827 and SCAFFOLD with a CVI of 0.1430 exhibit erratic performance. \sys{} benefits significantly from selecting the most relevant subset of experts for different settings.
\subsection{Ablation Study}
\begin{table}[t]
\centering
\caption{Ablation study under Standard-Hetero-T setting.}
\label{tab:ablation}
\begin{tabular}{@{}>{\centering\arraybackslash}p{1.7cm}%
    >{\centering\arraybackslash}p{0.5cm}%
    >{\centering\arraybackslash}p{0.5cm}%
    >{\centering\arraybackslash}p{0.5cm}%
    >{\centering\arraybackslash}p{1.5cm}%
    >{\centering\arraybackslash}p{1.2cm}@{}}
\toprule
\multirow{3}{*}{Method} & \multicolumn{3}{c}{Performance} & \multicolumn{2}{c}{Expert Num} \\
                & task-TC & task-RC & task-TS & \multirow{2}{*}{Avg.} & \multirow{2}{*}{Min/Max} \\
\midrule
FedMoE          & \textbf{94.76}           & \textbf{86.64}           & \textbf{16.92}           & \textbf{78 $\rightarrow$ 65} & \textbf{52/92}           \\
w/o stage1      & 92.68           & 83.90           & 14.50           & 96 $\rightarrow$ 104 & 94/118         \\
w/o stage2      & 92.97           & 86.43           & 16.69           & 78                 & 50/114          \\
\bottomrule
\end{tabular}
\end{table}
Table~\ref{tab:ablation} compares \sys{} to variants without either the first or second stage, showing task performance, the evolution of average expert numbers throughout training, and the maximum and minimum expert numbers of clients after training. The results indicate that both stages are vital for achieving strong performance, particularly the first stage where expert-activated information is collected. The second stage plays a pivotal role in adjusting the number of experts. When built upon the solid starting point provided by the first stage, it effectively prunes the redundant experts, thereby improving resource efficiency while maintaining strong performance. The two-stage training paradigm achieves an optimal balance between performance and resource consumption.
\section{Related Work}
\subsubsection{Personalized Federated Learning (PFL).}
With the growing concern for data privacy and model efficiency, PFL has recently garnered extensive attention~\cite{vanhaesebrouck2017decentralized,tan2022towards, sabah2023model}.
A common strategy is regularization, which adds penalty terms to the loss function or update direction to guide the training process~\cite{li2020federated, yao2020continual, karimireddy2020scaffold}.
However, it is sensitive to regularization parameters and difficult to generalize to complex scenarios.
Another approach is to build heterogeneous submodels for clients, utilizing techniques such as model distillation~\cite{ni2022federated, lin2020ensemble, zhu2022resilient}, pruning~\cite{ilhan2023scalefl}, or quantization~\cite{ozkara2021quped}, allowing knowledge transfer between global model and submodels. 
However, this approach depends on proxy data and may prolong the training process.
Parameter decoupling makes heterogeneous submodels more structured by separating the model into shared and personalized layers~\cite{arivazhagan2019federated, wu2021hierarchical, ma2022layer, mei2021fedvf}. However, designing the optimal layer combination for each client is challenging.
Some studies allow clients to maintain their original heterogeneous models. The server then decomposes, groups, and aggregates these models based on module similarity~\cite{wang2024towards, wang2023flexifed}. However, calculating modular similarity for large-scale models causes significant computational complexity.

\sys{} is one of the first PFL methods that supports LLMs with billion-scale parameters and can efficiently scale up, while previous methods are impractical for LLMs due to their complexity.

\subsubsection{Mixture of Experts (MoE) Optimizations.}
MoE has demonstrated effectiveness in handling complex centralized learning issues, including multi-task~\cite{chen2023adamv, chen2023mod}, multi-domain~\cite{zhang2024m3oe}, and multi-scenario~\cite{zou2022automatic}.
For example, AdaMV-MoE~\cite{chen2023adamv} exhibits enhanced performance in multi-task vision recognition by automatically adjusting the number of experts for each task.
Although some FL studies draw inspiration from the MoE concept, they arbitrarily treat the global and local models as individual experts and simply combine their outputs~\cite{zec2020specialized, guo2021pfl, bai2022multi, zhang2023fedbrain}. Experts at model granularity leads to limited flexibility and inadequate personalization. 

\sys{} is the first to integrate MoE into FL for a generic and versatile FedLLM, while previous works involve managing a collection of models.
\section{Conclusions}
\sys{} addresses the challenges of data heterogeneity in FL by employing models with MoE architecture. Through personalized sub-MoE construction, modular aggregation, and dynamic model adjustments, \sys{} enhances overall performance across diverse tasks, while significantly reducing memory footprint and network traffic. Empirical experiments demonstrate its effectiveness in cross-task scenarios.

\bibliography{aaai25}

\begin{thebibliography}{55}
\providecommand{\natexlab}[1]{#1}

\bibitem[{Arivazhagan et~al.(2019)Arivazhagan, Aggarwal, Singh, and Choudhary}]{arivazhagan2019federated}
Arivazhagan, M.~G.; Aggarwal, V.; Singh, A.~K.; and Choudhary, S. 2019.
\newblock Federated learning with personalization layers.
\newblock \emph{arXiv preprint arXiv:1912.00818}.

\bibitem[{Bai et~al.(2022)Bai, Zhang, Wang, Qin, and Zhang}]{bai2022multi}
Bai, T.; Zhang, Y.; Wang, Y.; Qin, Y.; and Zhang, F. 2022.
\newblock Multi-site MRI classification using Weighted federated learning based on Mixture of Experts domain adaptation.
\newblock In \emph{2022 IEEE International Conference on Bioinformatics and Biomedicine (BIBM)}, 916--921. IEEE.

\bibitem[{Bonawitz et~al.(2019)Bonawitz, Eichner, Grieskamp, Huba, Ingerman, Ivanov, Kiddon, Kone{\v{c}}n{\`y}, Mazzocchi, McMahan et~al.}]{bonawitz2019towards}
Bonawitz, K.; Eichner, H.; Grieskamp, W.; Huba, D.; Ingerman, A.; Ivanov, V.; Kiddon, C.; Kone{\v{c}}n{\`y}, J.; Mazzocchi, S.; McMahan, B.; et~al. 2019.
\newblock Towards federated learning at scale: System design.
\newblock \emph{Proceedings of machine learning and systems}, 1: 374--388.

\bibitem[{Cai et~al.(2023)Cai, Chen, Liu, Srinivasa, Lee, Kompella, and Wang}]{cai2023many}
Cai, R.; Chen, X.; Liu, S.; Srinivasa, J.; Lee, M.; Kompella, R.; and Wang, Z. 2023.
\newblock Many-task federated learning: A new problem setting and a simple baseline.
\newblock In \emph{Proceedings of the IEEE/CVF Conference on Computer Vision and Pattern Recognition}, 5037--5045.

\bibitem[{Chen et~al.(2023{\natexlab{a}})Chen, Chen, Du, Rashwan, Yang, Chen, Wang, and Li}]{chen2023adamv}
Chen, T.; Chen, X.; Du, X.; Rashwan, A.; Yang, F.; Chen, H.; Wang, Z.; and Li, Y. 2023{\natexlab{a}}.
\newblock Adamv-moe: Adaptive multi-task vision mixture-of-experts.
\newblock In \emph{Proceedings of the IEEE/CVF International Conference on Computer Vision}, 17346--17357.

\bibitem[{Chen et~al.(2023{\natexlab{b}})Chen, Shen, Ding, Chen, Zhao, Learned-Miller, and Gan}]{chen2023mod}
Chen, Z.; Shen, Y.; Ding, M.; Chen, Z.; Zhao, H.; Learned-Miller, E.~G.; and Gan, C. 2023{\natexlab{b}}.
\newblock Mod-squad: Designing mixtures of experts as modular multi-task learners.
\newblock In \emph{Proceedings of the IEEE/CVF Conference on Computer Vision and Pattern Recognition}, 11828--11837.

\bibitem[{Fedus, Dean, and Zoph(2022)}]{fedus2022review}
Fedus, W.; Dean, J.; and Zoph, B. 2022.
\newblock A review of sparse expert models in deep learning.
\newblock \emph{arXiv preprint arXiv:2209.01667}.

\bibitem[{Fedus, Zoph, and Shazeer(2022)}]{fedus2022switch}
Fedus, W.; Zoph, B.; and Shazeer, N. 2022.
\newblock Switch transformers: Scaling to trillion parameter models with simple and efficient sparsity.
\newblock \emph{Journal of Machine Learning Research}, 23(120): 1--39.

\bibitem[{Guo et~al.(2021)Guo, Mei, Xiao, and Wu}]{guo2021pfl}
Guo, B.; Mei, Y.; Xiao, D.; and Wu, W. 2021.
\newblock PFL-MoE: personalized federated learning based on mixture of experts.
\newblock In \emph{Web and Big Data: 5th International Joint Conference, APWeb-WAIM 2021, Guangzhou, China, August 23--25, 2021, Proceedings, Part I 5}, 480--486. Springer.

\bibitem[{Gupta et~al.(2022)Gupta, Mukherjee, Subudhi, Gonzalez, Jose, Awadallah, and Gao}]{gupta2022sparsely}
Gupta, S.; Mukherjee, S.; Subudhi, K.; Gonzalez, E.; Jose, D.; Awadallah, A.~H.; and Gao, J. 2022.
\newblock Sparsely activated mixture-of-experts are robust multi-task learners.
\newblock \emph{arXiv preprint arXiv:2204.07689}.

\bibitem[{Hu et~al.(2021)Hu, Shen, Wallis, Allen-Zhu, Li, Wang, Wang, and Chen}]{hu2021lora}
Hu, E.~J.; Shen, Y.; Wallis, P.; Allen-Zhu, Z.; Li, Y.; Wang, S.; Wang, L.; and Chen, W. 2021.
\newblock Lora: Low-rank adaptation of large language models.
\newblock \emph{arXiv preprint arXiv:2106.09685}.

\bibitem[{Ilhan, Su, and Liu(2023)}]{ilhan2023scalefl}
Ilhan, F.; Su, G.; and Liu, L. 2023.
\newblock Scalefl: Resource-adaptive federated learning with heterogeneous clients.
\newblock In \emph{Proceedings of the IEEE/CVF Conference on Computer Vision and Pattern Recognition}, 24532--24541.

\bibitem[{Jiang et~al.(2024)Jiang, Sablayrolles, Roux, Mensch, Savary, Bamford, Chaplot, Casas, Hanna, Bressand et~al.}]{jiang2024mixtral}
Jiang, A.~Q.; Sablayrolles, A.; Roux, A.; Mensch, A.; Savary, B.; Bamford, C.; Chaplot, D.~S.; Casas, D. d.~l.; Hanna, E.~B.; Bressand, F.; et~al. 2024.
\newblock Mixtral of experts.
\newblock \emph{arXiv preprint arXiv:2401.04088}.

\bibitem[{Kairouz et~al.(2021)Kairouz, McMahan, Avent, Bellet, Bennis, Bhagoji, Bonawitz, Charles, Cormode, Cummings et~al.}]{kairouz2021advances}
Kairouz, P.; McMahan, H.~B.; Avent, B.; Bellet, A.; Bennis, M.; Bhagoji, A.~N.; Bonawitz, K.; Charles, Z.; Cormode, G.; Cummings, R.; et~al. 2021.
\newblock Advances and open problems in federated learning.
\newblock \emph{Foundations and trends{\textregistered} in machine learning}, 14(1--2): 1--210.

\bibitem[{Kaplan et~al.(2020)Kaplan, McCandlish, Henighan, Brown, Chess, Child, Gray, Radford, Wu, and Amodei}]{kaplan2020scaling}
Kaplan, J.; McCandlish, S.; Henighan, T.; Brown, T.~B.; Chess, B.; Child, R.; Gray, S.; Radford, A.; Wu, J.; and Amodei, D. 2020.
\newblock Scaling laws for neural language models.
\newblock \emph{arXiv preprint arXiv:2001.08361}.

\bibitem[{Karimireddy et~al.(2020)Karimireddy, Kale, Mohri, Reddi, Stich, and Suresh}]{karimireddy2020scaffold}
Karimireddy, S.~P.; Kale, S.; Mohri, M.; Reddi, S.; Stich, S.; and Suresh, A.~T. 2020.
\newblock Scaffold: Stochastic controlled averaging for federated learning.
\newblock In \emph{International conference on machine learning}, 5132--5143. PMLR.

\bibitem[{Kim et~al.(2021)Kim, Awan, Muzio, Salinas, Lu, Hendy, Rajbhandari, He, and Awadalla}]{kim2021scalable}
Kim, Y.~J.; Awan, A.~A.; Muzio, A.; Salinas, A. F.~C.; Lu, L.; Hendy, A.; Rajbhandari, S.; He, Y.; and Awadalla, H.~H. 2021.
\newblock Scalable and efficient moe training for multitask multilingual models.
\newblock \emph{arXiv preprint arXiv:2109.10465}.

\bibitem[{Krajewski et~al.(2024)Krajewski, Ludziejewski, Adamczewski, Pi{\'o}ro, Krutul, Antoniak, Ciebiera, Kr{\'o}l, Odrzyg{\'o}{\'z}d{\'z}, Sankowski et~al.}]{krajewski2024scaling}
Krajewski, J.; Ludziejewski, J.; Adamczewski, K.; Pi{\'o}ro, M.; Krutul, M.; Antoniak, S.; Ciebiera, K.; Kr{\'o}l, K.; Odrzyg{\'o}{\'z}d{\'z}, T.; Sankowski, P.; et~al. 2024.
\newblock Scaling laws for fine-grained mixture of experts.
\newblock \emph{arXiv preprint arXiv:2402.07871}.

\bibitem[{Li et~al.(2020)Li, Sahu, Zaheer, Sanjabi, Talwalkar, and Smith}]{li2020federated}
Li, T.; Sahu, A.~K.; Zaheer, M.; Sanjabi, M.; Talwalkar, A.; and Smith, V. 2020.
\newblock Federated optimization in heterogeneous networks.
\newblock \emph{Proceedings of Machine learning and systems}, 2: 429--450.

\bibitem[{Lim et~al.(2020)Lim, Luong, Hoang, Jiao, Liang, Yang, Niyato, and Miao}]{lim2020federated}
Lim, W. Y.~B.; Luong, N.~C.; Hoang, D.~T.; Jiao, Y.; Liang, Y.-C.; Yang, Q.; Niyato, D.; and Miao, C. 2020.
\newblock Federated learning in mobile edge networks: A comprehensive survey.
\newblock \emph{IEEE communications surveys \& tutorials}, 22(3): 2031--2063.

\bibitem[{Lin et~al.(2021)Lin, He, Zeng, Wang, Huang, Dupuy, Gupta, Soltanolkotabi, Ren, and Avestimehr}]{lin2021fednlp}
Lin, B.~Y.; He, C.; Zeng, Z.; Wang, H.; Huang, Y.; Dupuy, C.; Gupta, R.; Soltanolkotabi, M.; Ren, X.; and Avestimehr, S. 2021.
\newblock Fednlp: Benchmarking federated learning methods for natural language processing tasks.
\newblock \emph{arXiv preprint arXiv:2104.08815}.

\bibitem[{Lin et~al.(2020)Lin, Kong, Stich, and Jaggi}]{lin2020ensemble}
Lin, T.; Kong, L.; Stich, S.~U.; and Jaggi, M. 2020.
\newblock Ensemble distillation for robust model fusion in federated learning.
\newblock \emph{Advances in neural information processing systems}, 33: 2351--2363.

\bibitem[{Liu et~al.(2024)Liu, Zhang, Li, Yan, Gao, Chen, Yuan, Huang, Sun, Gao et~al.}]{liu2024sora}
Liu, Y.; Zhang, K.; Li, Y.; Yan, Z.; Gao, C.; Chen, R.; Yuan, Z.; Huang, Y.; Sun, H.; Gao, J.; et~al. 2024.
\newblock Sora: A review on background, technology, limitations, and opportunities of large vision models.
\newblock \emph{arXiv preprint arXiv:2402.17177}.

\bibitem[{Lu et~al.(2024)Lu, Liu, Xu, Zhou, Huang, Zhang, Yan, and Li}]{lu2024not}
Lu, X.; Liu, Q.; Xu, Y.; Zhou, A.; Huang, S.; Zhang, B.; Yan, J.; and Li, H. 2024.
\newblock Not All Experts are Equal: Efficient Expert Pruning and Skipping for Mixture-of-Experts Large Language Models.
\newblock \emph{arXiv preprint arXiv:2402.14800}.

\bibitem[{Ma et~al.(2022)Ma, Zhang, Guo, and Xu}]{ma2022layer}
Ma, X.; Zhang, J.; Guo, S.; and Xu, W. 2022.
\newblock Layer-wised model aggregation for personalized federated learning.
\newblock In \emph{Proceedings of the IEEE/CVF conference on computer vision and pattern recognition}, 10092--10101.

\bibitem[{Mei et~al.(2021)Mei, Guo, Xiao, and Wu}]{mei2021fedvf}
Mei, Y.; Guo, B.; Xiao, D.; and Wu, W. 2021.
\newblock Fedvf: Personalized federated learning based on layer-wise parameter updates with variable frequency.
\newblock In \emph{2021 IEEE International Performance, Computing, and Communications Conference (IPCCC)}, 1--9. IEEE.

\bibitem[{Narayan, Cohen, and Lapata(2018)}]{narayan2018don}
Narayan, S.; Cohen, S.~B.; and Lapata, M. 2018.
\newblock Don't give me the details, just the summary! topic-aware convolutional neural networks for extreme summarization.
\newblock \emph{arXiv preprint arXiv:1808.08745}.

\bibitem[{Ni, Shen, and Zhao(2022)}]{ni2022federated}
Ni, X.; Shen, X.; and Zhao, H. 2022.
\newblock Federated optimization via knowledge codistillation.
\newblock \emph{Expert Systems with Applications}, 191: 116310.

\bibitem[{Ozkara et~al.(2021)Ozkara, Singh, Data, and Diggavi}]{ozkara2021quped}
Ozkara, K.; Singh, N.; Data, D.; and Diggavi, S. 2021.
\newblock Quped: Quantized personalization via distillation with applications to federated learning.
\newblock \emph{Advances in Neural Information Processing Systems}, 34: 3622--3634.

\bibitem[{Rajbhandari et~al.(2022)Rajbhandari, Li, Yao, Zhang, Aminabadi, Awan, Rasley, and He}]{rajbhandari2022deepspeed}
Rajbhandari, S.; Li, C.; Yao, Z.; Zhang, M.; Aminabadi, R.~Y.; Awan, A.~A.; Rasley, J.; and He, Y. 2022.
\newblock Deepspeed-moe: Advancing mixture-of-experts inference and training to power next-generation ai scale.
\newblock In \emph{International conference on machine learning}, 18332--18346. PMLR.

\bibitem[{Rajpurkar et~al.(2016)Rajpurkar, Zhang, Lopyrev, and Liang}]{rajpurkar2016squad}
Rajpurkar, P.; Zhang, J.; Lopyrev, K.; and Liang, P. 2016.
\newblock Squad: 100,000+ questions for machine comprehension of text.
\newblock \emph{arXiv preprint arXiv:1606.05250}.

\bibitem[{Sabah et~al.(2023)Sabah, Chen, Yang, Azam, Ahmad, and Sarwar}]{sabah2023model}
Sabah, F.; Chen, Y.; Yang, Z.; Azam, M.; Ahmad, N.; and Sarwar, R. 2023.
\newblock Model optimization techniques in personalized federated learning: A survey.
\newblock \emph{Expert Systems with Applications}, 122874.

\bibitem[{Shazeer et~al.(2017)Shazeer, Mirhoseini, Maziarz, Davis, Le, Hinton, and Dean}]{shazeer2017outrageously}
Shazeer, N.; Mirhoseini, A.; Maziarz, K.; Davis, A.; Le, Q.; Hinton, G.; and Dean, J. 2017.
\newblock Outrageously large neural networks: The sparsely-gated mixture-of-experts layer.
\newblock \emph{arXiv preprint arXiv:1701.06538}.

\bibitem[{Sun et~al.(2022)Sun, Yang, Liu, Yin, Li, and Xu}]{sun2022recent}
Sun, Z.; Yang, H.; Liu, K.; Yin, Z.; Li, Z.; and Xu, W. 2022.
\newblock Recent advances in LoRa: A comprehensive survey.
\newblock \emph{ACM Transactions on Sensor Networks}, 18(4): 1--44.

\bibitem[{Tan et~al.(2022)Tan, Yu, Cui, and Yang}]{tan2022towards}
Tan, A.~Z.; Yu, H.; Cui, L.; and Yang, Q. 2022.
\newblock Towards personalized federated learning.
\newblock \emph{IEEE transactions on neural networks and learning systems}, 34(12): 9587--9603.

\bibitem[{Vanhaesebrouck, Bellet, and Tommasi(2017)}]{vanhaesebrouck2017decentralized}
Vanhaesebrouck, P.; Bellet, A.; and Tommasi, M. 2017.
\newblock Decentralized collaborative learning of personalized models over networks.
\newblock In \emph{Artificial Intelligence and Statistics}, 509--517. PMLR.

\bibitem[{Villalobos et~al.(2022)Villalobos, Sevilla, Heim, Besiroglu, Hobbhahn, and Ho}]{villalobos2022will}
Villalobos, P.; Sevilla, J.; Heim, L.; Besiroglu, T.; Hobbhahn, M.; and Ho, A. 2022.
\newblock Will we run out of data? an analysis of the limits of scaling datasets in machine learning.
\newblock \emph{arXiv preprint arXiv:2211.04325}.

\bibitem[{Wang et~al.(2024)Wang, Yang, Cui, Che, Lyu, Xu, and Ma}]{wang2024towards}
Wang, J.; Yang, X.; Cui, S.; Che, L.; Lyu, L.; Xu, D.~D.; and Ma, F. 2024.
\newblock Towards personalized federated learning via heterogeneous model reassembly.
\newblock \emph{Advances in Neural Information Processing Systems}, 36.

\bibitem[{Wang et~al.(2023)Wang, He, Chen, Chen, Huang, Jin, and Yang}]{wang2023flexifed}
Wang, K.; He, Q.; Chen, F.; Chen, C.; Huang, F.; Jin, H.; and Yang, Y. 2023.
\newblock Flexifed: Personalized federated learning for edge clients with heterogeneous model architectures.
\newblock In \emph{Proceedings of the ACM Web Conference 2023}, 2979--2990.

\bibitem[{Wikipedia(2024)}]{wiki:GDPR}
Wikipedia. 2024.
\newblock General Data Protection Regulation --- Wikipedia{,} The Free Encyclopedia.
\newblock [Online; accessed 27-June-2024].

\bibitem[{{Wikipedia contributors}(2024)}]{wiki:CCPA}
{Wikipedia contributors}. 2024.
\newblock California Consumer Privacy Act --- {Wikipedia}{,} The Free Encyclopedia.
\newblock [Online; accessed 27-June-2024].

\bibitem[{Wolf et~al.(2019)Wolf, Debut, Sanh, Chaumond, Delangue, Moi, Cistac, Rault, Louf, Funtowicz et~al.}]{wolf2019huggingface}
Wolf, T.; Debut, L.; Sanh, V.; Chaumond, J.; Delangue, C.; Moi, A.; Cistac, P.; Rault, T.; Louf, R.; Funtowicz, M.; et~al. 2019.
\newblock Huggingface's transformers: State-of-the-art natural language processing.
\newblock \emph{arXiv preprint arXiv:1910.03771}.

\bibitem[{Wu et~al.(2021)Wu, Liu, Huang, Ning, Wang, Chen, Yi, and Zhou}]{wu2021hierarchical}
Wu, J.; Liu, Q.; Huang, Z.; Ning, Y.; Wang, H.; Chen, E.; Yi, J.; and Zhou, B. 2021.
\newblock Hierarchical personalized federated learning for user modeling.
\newblock In \emph{Proceedings of the Web Conference 2021}, 957--968.

\bibitem[{Wu et~al.(2023)Wu, He, Liu, Sun, Liu, Han, and Tang}]{wu2023brief}
Wu, T.; He, S.; Liu, J.; Sun, S.; Liu, K.; Han, Q.-L.; and Tang, Y. 2023.
\newblock A brief overview of ChatGPT: The history, status quo and potential future development.
\newblock \emph{IEEE/CAA Journal of Automatica Sinica}, 10(5): 1122--1136.

\bibitem[{Xu et~al.(2024)Xu, Cai, Wu, Li, and Wang}]{xu2024fwdllm}
Xu, M.; Cai, D.; Wu, Y.; Li, X.; and Wang, S. 2024.
\newblock $\{$FwdLLM$\}$: Efficient Federated Finetuning of Large Language Models with Perturbed Inferences.
\newblock In \emph{2024 USENIX Annual Technical Conference (USENIX ATC 24)}, 579--596.

\bibitem[{Yao et~al.(2022)Yao, Gao, Wang, Xie, Kuang, Chen, Wang, Dong, Ding, and Li}]{yao2022benchmark}
Yao, L.; Gao, D.; Wang, Z.; Xie, Y.; Kuang, W.; Chen, D.; Wang, H.; Dong, C.; Ding, B.; and Li, Y. 2022.
\newblock A benchmark for federated hetero-task learning.
\newblock \emph{arXiv preprint arXiv:2206.03436}.

\bibitem[{Yao and Sun(2020)}]{yao2020continual}
Yao, X.; and Sun, L. 2020.
\newblock Continual local training for better initialization of federated models.
\newblock In \emph{2020 IEEE International Conference on Image Processing (ICIP)}, 1736--1740. IEEE.

\bibitem[{Yi et~al.(2023)Yi, Guo, Wei, Zhou, Wang, and Xu}]{yi2023edgemoe}
Yi, R.; Guo, L.; Wei, S.; Zhou, A.; Wang, S.; and Xu, M. 2023.
\newblock Edgemoe: Fast on-device inference of moe-based large language models.
\newblock \emph{arXiv preprint arXiv:2308.14352}.

\bibitem[{Zec et~al.(2020)Zec, Mogren, Martinsson, S{\"u}tfeld, and Gillblad}]{zec2020specialized}
Zec, E.~L.; Mogren, O.; Martinsson, J.; S{\"u}tfeld, L.~R.; and Gillblad, D. 2020.
\newblock Specialized federated learning using a mixture of experts.
\newblock \emph{arXiv preprint arXiv:2010.02056}.

\bibitem[{Zhang et~al.(2023)Zhang, Meng, Liu, Wu, Wang, and Ning}]{zhang2023fedbrain}
Zhang, C.; Meng, X.; Liu, Q.; Wu, S.; Wang, L.; and Ning, H. 2023.
\newblock FedBrain: A robust multi-site brain network analysis framework based on federated learning for brain disease diagnosis.
\newblock \emph{Neurocomputing}, 559: 126791.

\bibitem[{Zhang, Zhao, and LeCun(2015)}]{zhang2015character}
Zhang, X.; Zhao, J.; and LeCun, Y. 2015.
\newblock Character-level convolutional networks for text classification.
\newblock \emph{Advances in neural information processing systems}, 28.

\bibitem[{Zhang et~al.(2024)Zhang, Liu, Yu, Cai, Zhao, Zhang, Liu, Liu, Zhao, Hu et~al.}]{zhang2024m3oe}
Zhang, Z.; Liu, S.; Yu, J.; Cai, Q.; Zhao, X.; Zhang, C.; Liu, Z.; Liu, Q.; Zhao, H.; Hu, L.; et~al. 2024.
\newblock M3oE: Multi-Domain Multi-Task Mixture-of Experts Recommendation Framework.
\newblock \emph{arXiv preprint arXiv:2404.18465}.

\bibitem[{Zhu et~al.(2021)Zhu, Xu, Liu, and Jin}]{zhu2021federated}
Zhu, H.; Xu, J.; Liu, S.; and Jin, Y. 2021.
\newblock Federated learning on non-IID data: A survey.
\newblock \emph{Neurocomputing}, 465: 371--390.

\bibitem[{Zhu et~al.(2022)Zhu, Hong, Drew, and Zhou}]{zhu2022resilient}
Zhu, Z.; Hong, J.; Drew, S.; and Zhou, J. 2022.
\newblock Resilient and communication efficient learning for heterogeneous federated systems.
\newblock \emph{Proceedings of machine learning research}, 162: 27504.

\bibitem[{Zou et~al.(2022)Zou, Hu, Zhao, Ding, Liu, Li, and Sun}]{zou2022automatic}
Zou, X.; Hu, Z.; Zhao, Y.; Ding, X.; Liu, Z.; Li, C.; and Sun, A. 2022.
\newblock Automatic expert selection for multi-scenario and multi-task search.
\newblock In \emph{Proceedings of the 45th International ACM SIGIR Conference on Research and Development in Information Retrieval}, 1535--1544.

\end{thebibliography}

\end{document}